\documentclass[runningheads]{llncs}
\usepackage[T1]{fontenc}
\let\llncssubparagraph\subparagraph
\let\subparagraph\paragraph
\usepackage[compact]{titlesec}
\let\subparagraph\llncssubparagraph
\usepackage{array}
\newcolumntype{P}[1]{>{\centering\arraybackslash}p{#1}}
\usepackage{float}
\usepackage{array}
\usepackage{tabularx}
\usepackage{cite}
\usepackage{graphicx}
\usepackage{amsmath}
\usepackage{multirow}
\usepackage{booktabs}
\usepackage[bb=px]{mathalfa} 
\usepackage{adjustbox}
\usepackage{filecontents}
\usepackage{lipsum}
\usepackage{fancyhdr}
\usepackage{makecell} 
\expandafter\def\expandafter\normalsize\expandafter{%
    \normalsize%
    \setlength\abovedisplayskip{0pt}%
    \setlength\belowdisplayskip{8pt}%
    \setlength\abovedisplayshortskip{-8pt}%
    \setlength\belowdisplayshortskip{2pt}%
}
\begin{document}
%
% \title{Leveraging Masked Autoencoder to Improve Transformer-based Pain Assessment}
\title{Transformer with Leveraged Masked Autoencoder for video-based Pain Assessment}
% \title{Learning Sequential Contexts using Latent Diffusion for Dyadic Reaction Generation\thanks{Supported by BRL \& IITP.}}
% \author{Minh-Duc Nguyen\inst{1}\orcidID{0000-1111-2222-3333} \and
% Hyung-Jeong Yang\inst{2,3}\orcidID{1111-2222-3333-4444} \and
% Third Author\inst{3}\orcidID{2222--3333-4444-5555}\and
% Third Author\inst{3}\orcidID{2222--3333-4444-5555}\and
% Third Author\inst{3}\orcidID{2222--3333-4444-5555}
% }
\author{Minh-Duc Nguyen \and
Hyung-Jeong Yang\inst{*}\and
Soo-Hyung Kim \and
Ji-Eun Shin\and
Seung-Won Kim
}
%
% \authorrunning{Nguyen et al.}
\institute{Chonnam National University,
Gwangju, South Korea \\
% *Corresponding author e-mail: hjyang@jnu.ac.kr}
% \url{http://www.springer.com/gp/computer-science/lncs}\and
{\{ducnm,hjyang,shkim,jieunshin,seungwon.kim\}@jnu.ac.kr}}

% \email{\{dunm,hjyang,shkim,jieunshin,seungwon.kim\}@jnu.ac.kr}}
%
\titlerunning{Latent Behavior Diffusion for Sequential Reaction Generation}
\pagestyle{fancy}
\fancyhf{} % Clear all header and footer fields
\fancypagestyle{plain}{ % For plain pages
  \fancyhf{} % Clear all header and footer fields
}
\renewcommand{\headrulewidth}{0pt} % Remove the header rule line
\maketitle              % typeset the header of the contribution
\renewcommand{\thefootnote}{}
\footnote{* Corresponding author.} 
\vspace{-1em}
\begin{abstract}
Accurate pain assessment is crucial in healthcare for effective diagnosis and treatment; however, traditional methods relying on self-reporting are inadequate for populations unable to communicate their pain. Cutting-edge AI is promising for supporting clinicians in pain recognition using facial video data. In this paper, we enhance pain recognition by employing facial video analysis within a Transformer-based deep learning model. By combining a powerful Masked Autoencoder with a Transformers-based classifier, our model effectively captures pain level indicators through both expressions and micro-expressions. We conducted our experiment on the AI4Pain dataset, which produced promising results that pave the way for innovative healthcare solutions that are both comprehensive and objective.

\end{abstract}
\titlespacing*{\section}{0pt}{0.2\baselineskip}{0.2\baselineskip}
\section{Introduction}

Pain assessment is a critical component of healthcare and is essential for accurate diagnosis, effective treatment, and overall patient well-being. However, the inherently subjective nature of pain presents significant challenges, particularly in populations unable to communicate their discomfort effectively, such as infants, non-verbal patients, and those with cognitive impairments. The burgeoning affective computing field offers promising advancements by integrating artificial intelligence (AI) and sophisticated sensing technologies. This interdisciplinary effort explores the potential of facial video analysis to enhance pain assessment accuracy and depth, ultimately contributing to improved patient care and more empathetic clinical interventions.

Traditional methods of pain assessment predominantly rely on self-reporting because pain is a subjective experience. However, self-reporting is not always valid and trustworthy, for example, in individuals with mental illnesses  \cite{p1}. Moreover, it is inapplicable to patients who are unconscious or are infants. Physiological and observational measurements \cite{p2} can be useful in certain situations. 
% They could also make it easier to overcome the drawbacks of basic rating scales, which are frequently used in clinics \cite{p3}. 
For proper pain treatment, the evaluation must be performed frequently, especially if the patient is unable to ask for aid; therefore, it is necessary to develop more objective and comprehensive assessment tools.

In this paper, we explore the potential of facial video analysis to enhance pain assessment accuracy and depth, ultimately contributing to improved patient care and more empathetic clinical interventions. We leverage a Transformer-based architecture that contains: An effective Masked Autoencoder for temporal feature extraction, based on the state-of-the-art MARLIN \cite{Marlin}
and a Transformer-based classifier for Multivariate Time Series Classification.
We applied the two position encoding techniques: time Absolute Position Encoding (tAPE) and efficient Relative Position Encoding (eRPE) from \cite{convTrans}.

\titlespacing*{\section}{0pt}{0.5\baselineskip}{0.2\baselineskip}
\section{Related Works}
Deep neural networks (DNNs) have been increasingly applied to pain assessment, addressing the need for more accurate and objective methods. Among these approaches, DNN-based models are designed to analyze pain-induced facial expressions. These models are characterized by their ability to extract relevant descriptors and optimize neural network-based inference models directly from processed raw input data. In \cite{ref3,ref4}, the authors presented a hybrid deep neural network architecture for pain detection. This architecture integrates a feature embedding network comprising a Convolutional Neural Network (CNN) and a Long Short-Term Memory (LSTM \cite{lstm}), leveraging both the spatial and temporal characteristics of facial pain expressions in video sequences. Some previous work \cite{ref5,ref6} employed a transfer learning approach, utilizing deep learning models that were pre-trained not on pain assessment but on different tasks such as face recognition and object classification.

Over the past few years, advancements in deep-learning techniques have significantly enhanced the capabilities of pain recognition tasks. Lu et al. \cite{ref1} introduce a Cross-Stream Attention (CSA) mechanism utilizing non-local operations to capture the correlation between two Convolutional Network streams. This approach allows for spatial and temporal information interactions at various semantic levels, specifically tailored for analyzing neonatal pain expression videos. A multimodal automatic assessment framework for acute pain that utilizes both video and heart rate signals was also introduced in \cite{ref7}. 

Most deep-learning approaches use CNN or a pretraining such as VGGFace \cite{vggface} to extract facial features for video data that relies on local convolutions that focus more on local feature extraction. Vision Transformers (ViT) \cite{vit} excel at capturing global context and long-range dependencies in an image through their self-attention mechanism, and often perform better with larger datasets and more computational resources due to their ability to scale effectively. 
ViT also can learn more complex and nuanced representations of images because they consider the relationships between all parts of the image simultaneously. We applied transfer learning to a very deep masked auto-encoder, MARLIN \cite{Marlin} which adopted ViT architecture as a backbone, to extract spatio-temporal features. Subsequently, a Convolutional Transformer with skip connections was employed to predict pain levels using features extracted by MARLIN encoder.

% In this study, we proposed a Transformer-based approach for deep-face analysis to tackle pain assessment tasks. We applied transfer learning to a very deep masked auto-encoder, MARLIN \cite{Marlin} which adopted ViT architecture as a backbone, to extract spatio-temporal features. Subsequently, a Convolutional Transformer with skip connections was employed to predict pain levels using features extracted by MARLIN. Two key ideas from \cite{convTrans}, namely tAPE and eRPE, were adapted to enhance the position encoding of Attention module for classification tasks effectively.
% \subsection{Diffusion Generative Models}
% Diffusion Generative Models (DGMs) use an iterative process of adding noise to an initial input to generate high-quality samples, popular for their ability to produce realistic images and sequences.

\titlespacing*{\section}{0pt}{0.3\baselineskip}{0.2\baselineskip}

\section{Proposal Method}
Our proposed model combines a Transformer-based classifier with a Facial Encoder head to address the pain assessment problem. 
% The spatio-temporal feature extraction retains the competency of the power MARLIN encoder and at the same time, the Transformer model can learn the context. 
Given a sequence of frames from a video clip, segmented by a window size $L$, the facial feature extractor takes advantage of the MARLIN encoder. Then it creates a series of output vectors, each representing the extracted features from a single frame. These output vectors serve as the input to our Residual ConvTrans network to predict the corresponding pain level.
% \vspace{-2em}
\begin{figure}[t!]
\centerline{\includegraphics[scale=0.5]{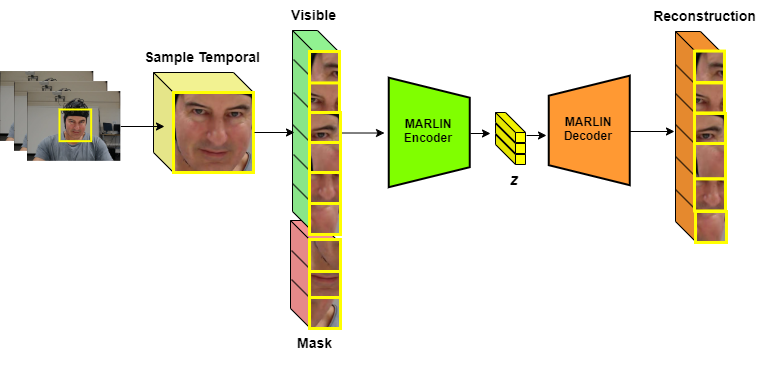}}
\caption{MARLIN Pre-training \cite{Marlin}.}
\label{marlin}
\end{figure}
% \vspace{-2em}
\titlespacing*{\subsection}{0pt}{0.3\baselineskip}{0.2\baselineskip}
\subsection{Spatio-temporal Feature Compression}\label{AA}
We leverage the MARLIN encoder as our facial feature extractor to encode the face structure and texture, which carry important indications regarding an individual’s perceived emotion. Based on ViT architecture, MARLIN stands as a pre-trained transformer model that has undergone extensive learning, primarily focusing on densely masked facial regions such as the eyes, nose, mouth, lips, and skin. Its training revolves around reconstructing the intricate spatio-temporal characteristics observed in facial videos. Through this reconstruction process, MARLIN captures both local and global features inherent to facial expressions. This capability enables it to encode a wide array of general and transferable facial attributes across various tasks. 
% Notably, MARLIN’s training data comprise large-scale web-crawled videos, and it utilizes self-supervised learning methods. This approach ensures MARLIN’s robustness in learning facial video features, empowering it with the ability to extract meaningful representations from facial data effectively. 
We first use MARLIN pre-training to fine-tune the whole self-supervised Masked Auto Encoder with AI4Pain video dataset
% , which contains a total of 65 participants samples in three categories: No Pain (NP), Low Pain (Low), and High Pain (High). 
The pre-training stage can be shown in Fig. \ref{marlin}, the model is optimized by $L2$ reconstruction loss with given an input masked tokens $\tilde{X}_{m}$, the masked auto-encoder module reconstruct it back to $X_{m}^{\prime}$:

\begin{equation*}
\mathcal{L}_{\text {rec }}=\frac{1}{N} \sum_{i=1}^{N}\left\|X_{m}^{(i)}-X_{m}^{\prime(i)}\right\|_{2} \tag{1}
\end{equation*}
MARLIN encoder is used as a powerful tool for extracting meaningful facial features from segmented frames. 
% It operates on each segment individually to extract the relevant facial characteristics. 
The output of the MARLIN encoder is a series of output vectors, each representing the extracted features from a single frame. These vectors encapsulate essential facial information present in each segment.

\subsection{Residual Transformer Classifier}
We utilize a Transformer-based architecture to capture the sequential nature of time series as the facial spatio-temporal features. Followed by \cite{convTrans}, we combined tAPE and eRPE position encoding techniques into a single framework by employing their introduced ConvTrans model.
\titlespacing*{\subsubsection}{0pt}{0.2\baselineskip}{0.2\baselineskip}
\subsubsection{Absolute position encoding.}

The original self-attention mechanism, as described by Vaswani et al. \cite{trans}, incorporates absolute positional embeddings to the input sequence.  The positional embedding $P$ is added to the input embedding $x$. Fixed position encodings using sine and cosine functions are proposed to provide the model with information about the relative and absolute position of the tokens in the sequence with length $L$. The encoding method, tAPE, $\omega_{k}^{\text {new }}$, takes into account both the input embedding dimension and the length of the sequence, enhancing the model's ability to capture temporal dependencies.
\begin{gather*}
    x_{i}=x_{i}+p_{i} \tag{2}\\
 p_{i}(2 k)=\sin i \omega_{k} \tag{3}\\
 p_{i}(2 k+1)=\cos i \omega_{k} \tag{4}\\
\omega_{k}=10000^{-2 k / d_{\text {model }}} \tag{5}\\
 \omega_{k}^{\text {new }}=\frac{\omega_{k} \times d_{\text {model }}}{L} \tag{6}
\end{gather*}

where the position embedding $p_{i} \in \mathbb{R}^{d_{\text {model }}}$, $k$ is in the range of $\left[0, \frac{d_{\text {model }}}{2}\right]$, $d_{\text {model }}$ is the embedding dimension and $\omega_{k}$ is the frequency term.

\subsubsection{Relative position encoding.}

% To implement the efficient version of eRFE for input time series with a length of $L$
% , for each head, we create a trainable parameter $w$
%  of size $2L-1$
% , as the maximum distance is $2L-1$
% . Then for two position indices $i$
%  and $j$
% , the corresponding relative scalar is $w_{i-j+L}$
%  where indexes start from 1 instead of 0 (1-base index). Accordingly, we need to index $L^2$ 
%  elements from $2L-1$
%  vector.
In Transformers, a query and a set of key-value pairs are used to produce an output. Specifically, for an input sequence $\mathbf{x}_{\mathbf{t}}=\left\{x_{1}, x_{2}, \ldots, x_{L}\right\}$
, self-attention calculates an output sequence $\mathbf{z}_{\mathbf{t}}$=$\left\{z_{1}, z_{2}, \ldots, z_{L}\right\}$
, where each $z_{i}$
  is a vector in  $\mathbb{R}^{d_{z}}$
. Each $z_{i}$ is derived as a weighted sum of the input elements.

\begin{equation*}
z_{i}=\sum_{j=1}^{L} \alpha_{i, j}\left(x_{j} W^{V}\right) \tag{7}
\end{equation*}

Each coefficient weight $\alpha_{i, j}$ is calculated using softmax function:

\begin{equation*}
\alpha_{i, j}=\frac{\exp \left(e_{i j}\right)}{\sum_{k=1}^{L} \exp \left(e_{i k}\right)} \tag{8}
\end{equation*}
where $e_{i j}$ is an attention weight from positions $j$ to $i$ and is computed using a scaled  dot-product. The eRPE enchanted the self-attention module by the following formula:

\begin{equation*}
\alpha_{i}=\sum_{j \in L}(\underbrace{\frac{\exp \left(e_{i, j}\right)}{\sum_{k \in L} \exp \left(e_{i, k}\right)}}_{A_{i, j}}+w_{i-j}) x_{j} \tag{9}
\end{equation*}

where $L$ is series length, $A_{i, j}$ is attention weight and $w_{i-j}$ is a learnable scalar (i.e., $\left.w \in \mathbb{R}^{O(L)}\right)$ and represent the relative position weight between positions $i$ and $j$.

For time series data of length  $L$, each attention head uses a trainable parameter 
$w$
 of size 
$2L-1$, which accommodates the maximum relative distance. For any two positions 
$i$ and 
$j$, the relative scalar is determined by 
$w_{i-j+L}$, with indices starting from 1. This approach necessitates indexing 
$L^2$  elements from the 
 $2L-1$ vector to cover all possible position pairs in the sequence. This method ensures that the model effectively captures the relative positional information, enhancing its ability to model temporal dependencies and patterns.

\subsubsection{Convolutional Transformer Model.}
ConvTrans, introduced by \cite{convTrans}, takes tAPE-generated position embeddings that are added to ensure that the model captures the temporal order in time series data before input embeddings enter the transformer block. Subsequently, attention is performed within the Multi-Head attention block by eRPE. FFN is a multi-layer perceptron block consisting of two linear layers and Gaussian Error Linear Units (GELUs) as an activation function. We leverage ConvTrans as our classifier for Pain assessment.
% \vspace{-1em}
\begin{figure*}[t!]
\centerline{\includegraphics[scale=0.2]{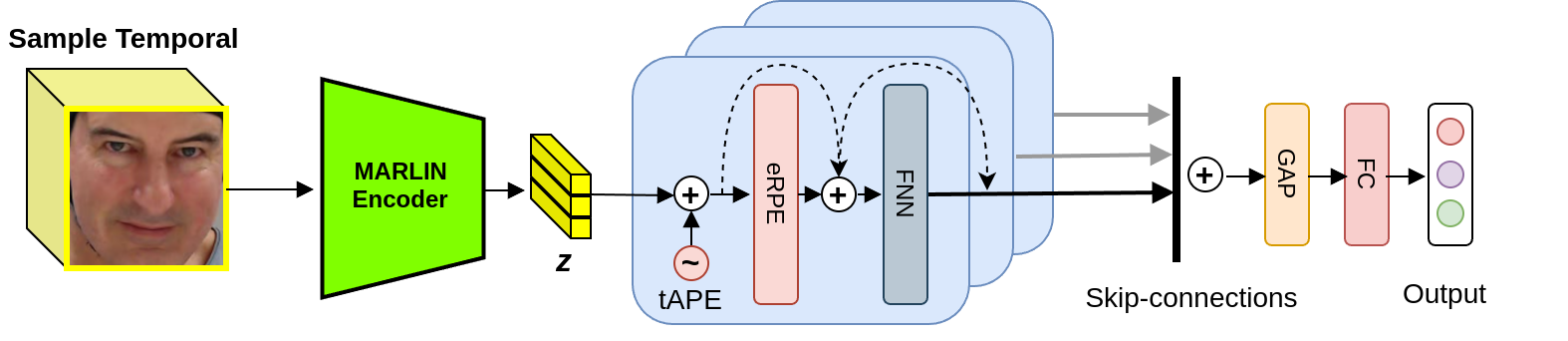}}
\caption{Overall of Proposed architecture}
\label{dual}
\end{figure*}
\titlespacing*{\subsubsection}{0pt}{0.1\baselineskip}{0.2\baselineskip}

% \vspace{-1em}
\subsubsection{Framework Architecture.} Our unified framework as depicted in Fig. \ref{dual}. First, MARLIN e takes a sliding window's segmented sequence of frames and creates fine-grained spatio-temporal features. A sequence of residual ConvTrans layers with skip connections is applied to capture MARLIN features as Multivariate Time Series input and then performs its Transformer block. After obtaining the final output from the transformer block, max-pooling and global average pooling (GAP) are applied to the output of the last layer’s ELU activation function, enhancing the model’s translation invariance. The loss function is calculated as Cross Entropy Loss: 

\begin{equation*}
    L_{CE}=\sum_{c=1}^My_{o,c}\log(p_{o,c}) \tag{10}
\end{equation*}

where $M$ is the number of classes, $y$ is the binary indicator (0 or 1) if class label $c$ is the correct classification for observation $o$ and $p$ indicates predicted probability observation $o$ is of class $c$.
\section{Experiments}
\subsubsection{Dataset.}
 % We use the grand challenge dataset for pain assessment AI4Pain \cite{fernandez2023multimodal}. 
 % % The methodology employed for data collection closely adhered to the approach detailed in our previous study. 
 % The dataset, collected at the Human-Machine Interface Laboratory at the University of Canberra, includes multimodal recordings of participants’ pain responses. Participants were seated with both arms resting on a table, and data were gathered using a functional near-infrared spectroscopy (fNIRS) headset positioned on the frontal area and a Logitech StreamCam capturing facial video at 30 Hz. Pain was induced via a transcutaneous electrical nerve stimulation (TENS) machine with electrodes on the right arm. The experiment involved a  60-second baseline recording and six repetitions of low and high pain stimuli at two locations, counterbalanced to avoid habituation and order effects, resulting in 12 repetitions per pain condition. The dataset focuses on neural activity and facial expressions, providing valuable data for pain assessment research. In this study, we only use our model on video modality. By applying off-the-shelf face SDK modules with a pre-trained face detection model from \cite{face} to crop raw videos into facial sequence images with the size of 224x224. The objective is to classify sample data into one of three categories: No Pain (NP), Low Pain (Low), and High Pain (High).
 We use the grand challenge dataset for pain assessment AI4Pain \cite{fernandez2023multimodal}. 
 % The methodology employed for data collection closely adhered to the approach detailed in our previous study. 
 The dataset, collected at the Human-Machine Interface Laboratory at the University of Canberra, includes multimodal recordings of 65 participants’ pain responses: videos and functional near-infrared spectroscopy (fNIRS) data.
 % Participants were seated with both arms resting on a table, and data were gathered using a functional near-infrared spectroscopy (fNIRS) headset positioned on the frontal area and a Logitech StreamCam capturing facial video at 30 Hz. Pain was induced via a transcutaneous electrical nerve stimulation (TENS) machine with electrodes on the right arm. The experiment involved a  60-second baseline recording and six repetitions of low and high pain stimuli at two locations, counterbalanced to avoid habituation and order effects, resulting in 12 repetitions per pain condition. 
 The dataset focuses on neural activity and facial expressions, providing valuable data for pain assessment research. In this study, we only use our model on video modality. By applying off-the-shelf face SDK modules with a pre-trained face detection model from \cite{face} to crop raw videos into facial sequence images with the size of 224x224. The objective is to classify sample data into one of three categories: No Pain (NP), Low Pain (Low), and High Pain (High). 

\subsubsection{Experiment Settings. }
We implemented the method on PyTorch with an RTX 8000. For self-supervised pre-training, we used the AdamW optimizer with a base learning rate of $1.5e-4$ 
, momentum parameters 
$\beta_{1}=0.9$ and 
$\beta_{3}=0.95$, and a cosine decay learning rate scheduler,  the masking ratio is set to 0.9. For linear probing, we employed the Adam optimizer with 
$\beta_{1}=0.5$, 
$\beta_{2}=0.9$, a base learning rate of $1e-4$. In the second stage, we train ConvTrans + Skip with 8 layers with MARLIN encoder head. The segment length is set to $L$=16, and the dimension of the dense feedforward part of the transformer layer is 256 with 8 attention heads, we applied learning rate of $1e-3$ with RAdam optimizer.
% \vspace{-1em}
\begin{table}[t!]
\centering
\caption{The results of various methods on the AI4PAIN validation set.}
\label{tab:1}
\resizebox{\columnwidth}{!}
{%
\begin{tabular}{|c|llll|llll|llll|l|}

\Xhline{1.5pt}  
\multirow{2}{*}{Models} & \multicolumn{4}{c|}{Precision}                                                                            & \multicolumn{4}{c|}{Recall}                                                                               & \multicolumn{4}{c|}{F1-score}                                                                             & \multicolumn{1}{c|}{\multirow{2}{*}{Acc.}} \\ \cline{2-13}
                        & \multicolumn{1}{c|}{NP} & \multicolumn{1}{c|}{Low} & \multicolumn{1}{c|}{High} & \multicolumn{1}{c|}{Avg} & \multicolumn{1}{c|}{NP} & \multicolumn{1}{c|}{Low} & \multicolumn{1}{c|}{High} & \multicolumn{1}{c|}{Avg} & \multicolumn{1}{c|}{NP} & \multicolumn{1}{c|}{Low} & \multicolumn{1}{c|}{High} & \multicolumn{1}{c|}{Avg} & \multicolumn{1}{c|}{}                      \\ \hline
\midrule
PyFeat+Gaussian SVM   & \multicolumn{1}{l|}{\hfil -}   & \multicolumn{1}{l|}{\hfil -}    & \multicolumn{1}{l|}{\hfil -}     &            {\hfil -}              & \multicolumn{1}{l|}{\hfil -}   & \multicolumn{1}{l|}{\hfil -}    & \multicolumn{1}{l|}{\hfil -}     &           {\hfil -}               & \multicolumn{1}{l|}{\hfil -}   & \multicolumn{1}{l|}{\hfil -}    & \multicolumn{1}{l|}{\hfil -}     &        {\hfil -}                  &   0.40                                         \\ 
% Baseline (fNIRS)        & \multicolumn{1}{l|}{}   & \multicolumn{1}{l|}{}    & \multicolumn{1}{l|}{}     &                          & \multicolumn{1}{l|}{}   & \multicolumn{1}{l|}{}    & \multicolumn{1}{l|}{}     &                          & \multicolumn{1}{l|}{}   & \multicolumn{1}{l|}{}    & \multicolumn{1}{l|}{}     &                          &   0.40                                         \\ 
% Baseline (Video+fNIRS)        & \multicolumn{1}{l|}{}   & \multicolumn{1}{l|}{}    & \multicolumn{1}{l|}{}     &                          & \multicolumn{1}{l|}{}   & \multicolumn{1}{l|}{}    & \multicolumn{1}{l|}{}     &                          & \multicolumn{1}{l|}{}   & \multicolumn{1}{l|}{}    & \multicolumn{1}{l|}{}     &                          &   0.40                                         \\ 
Twins-PainViT \cite{painvit}
& \multicolumn{1}{l|}{\hfil -}   & \multicolumn{1}{l|}{\hfil -}   & \multicolumn{1}{l|}{\hfil -}     &           {\hfil -}               & \multicolumn{1}{l|}{\hfil -}   & \multicolumn{1}{l|}{\hfil -}   & \multicolumn{1}{l|}{\hfil -}    &         {\hfil -}                 & \multicolumn{1}{l|}{\hfil -}   & \multicolumn{1}{l|}{\hfil -}    & \multicolumn{1}{l|}{\hfil -}     &         {\hfil -}                 &    0.45                                       \\ 
Simple ANN+Voting \cite{ai4pain}    & \multicolumn{1}{l|}{0.10}   & \multicolumn{1}{l|}{0.60}    & \multicolumn{1}{l|}{0.66}     &     0.45                     & \multicolumn{1}{l|}{0.17}   & \multicolumn{1}{l|}{0.67}    & \multicolumn{1}{l|}{0.55}     &         0.46                 & \multicolumn{1}{l|}{0.12}   & \multicolumn{1}{l|}{0.63}    & \multicolumn{1}{l|}{0.60}     &          0.45                &                0.59                            \\ 
VGG19+LSTM  \cite{ai4pain}                  & \multicolumn{1}{l|}{0.24}   & \multicolumn{1}{l|}{0.59}    & \multicolumn{1}{l|}{\textbf{0.71}}     &        0.51                  & \multicolumn{1}{l|}{0.42}   & \multicolumn{1}{l|}{\textbf{0.74}}    & \multicolumn{1}{l|}{0.49}     &         0.55                 & \multicolumn{1}{l|}{0.30}   & \multicolumn{1}{l|}{\textbf{0.65}}    & \multicolumn{1}{l|}{0.58}     &      0.51                    &                      0.60                      \\ 
Ours                    & \multicolumn{1}{l|}{\textbf{0.95}}   & \multicolumn{1}{l|}{\textbf{0.62}}   & \multicolumn{1}{l|}{0.60}     &        \textbf{0.72}                  & \multicolumn{1}{l|}{\textbf{1.00}}   & \multicolumn{1}{l|}{0.52}    & \multicolumn{1}{l|}{\textbf{0.63}}     &       \textbf{0.72}                  & \multicolumn{1}{l|}{\textbf{0.98}}   & \multicolumn{1}{l|}{0.56}   & \multicolumn{1}{l|}{\textbf{0.61}}     &            \textbf{0.72}             &              \textbf{0.79}                              \\ \Xhline{1.5pt}  
\end{tabular}%
}
\end{table}
\begin{table}[t!]
\centering
\caption{The results of various methods on the AI4PAIN test set.}
\label{tab:3}
\begin{tabular}{|c|c|}
\Xhline{1.5pt}  

Models      & Accuracy \\ \hline\hline
% Gaussian SVM (fNIRS)&      43.3    \\ \hline
Pyfeat + GaussianSVM &       0.40  \\ 
VGG19 + LSTM \cite{ai4pain}&     0.43     \\ 
Simple ANN + Voting \cite{ai4pain}&     0.49     \\ 
Marlin + Transformer&     0.52     \\ 
Ours             &      \textbf{0.55}    \\ \Xhline{1.5pt}  
\end{tabular}
\end{table}
% \vspace{-3em}
\begin{table}[t!]
\centering
\caption{Experiments on the AI4PAIN validation set with different classifiers.}
\label{tab:2}
\resizebox{\columnwidth}{!}{%
\begin{tabular}{|c|llll|llll|llll|l|}
\Xhline{1.5pt}  
\multirow{2}{*}{Models} & \multicolumn{4}{c|}{Precision}                                                                            & \multicolumn{4}{c|}{Recall}                                                                               & \multicolumn{4}{c|}{F1-score}                                                                             & \multicolumn{1}{c|}{\multirow{2}{*}{Acc.}} \\ \cline{2-13}
                        & \multicolumn{1}{c|}{NP} & \multicolumn{1}{c|}{Low} & \multicolumn{1}{c|}{High} & \multicolumn{1}{c|}{Avg} & \multicolumn{1}{c|}{NP} & \multicolumn{1}{c|}{Low} & \multicolumn{1}{c|}{High} & \multicolumn{1}{c|}{Avg} & \multicolumn{1}{c|}{NP} & \multicolumn{1}{c|}{Low} & \multicolumn{1}{c|}{High} & \multicolumn{1}{c|}{Avg} & \multicolumn{1}{c|}{}                      \\ \hline
\midrule
% Marlin+FCN        & \multicolumn{1}{l|}{}   & \multicolumn{1}{l|}{}    & \multicolumn{1}{l|}{}     &                          & \multicolumn{1}{l|}{}   & \multicolumn{1}{l|}{}    & \multicolumn{1}{l|}{}     &                          & \multicolumn{1}{l|}{}   & \multicolumn{1}{l|}{}    & \multicolumn{1}{l|}{}     &                          &                                            \\ 
Marlin+Transformer      & \multicolumn{1}{l|}{0.90}   & \multicolumn{1}{l|}{0.62}    & \multicolumn{1}{l|}{0.59}     &     0.74                     & \multicolumn{1}{l|}{0.99}   & \multicolumn{1}{l|}{0.47}    & \multicolumn{1}{l|}{0.61}     &           0.69               & \multicolumn{1}{l|}{0.94}   & \multicolumn{1}{l|}{0.54}    & \multicolumn{1}{l|}{0.60}     &         0.69                 &   0.76                                        \\ 
Marlin+LSTM                    & \multicolumn{1}{l|}{0.94}   & \multicolumn{1}{l|}{0.60}    & \multicolumn{1}{l|}{\textbf{0.61}  }   &          0.72                & \multicolumn{1}{l|}{1.00}   & \multicolumn{1}{l|}{\textbf{0.60}}    & \multicolumn{1}{l|}{0.54}     &       0.71                   & \multicolumn{1}{l|}{0.97}   & \multicolumn{1}{l|}{\textbf{0.60}}    & \multicolumn{1}{l|}{0.57}     &         0.71                 &                     0.78                       \\ 
Marlin+Res-ConvTrans                     & \multicolumn{1}{l|}{\textbf{0.95}}   & \multicolumn{1}{l|}{\textbf{0.62}}   & \multicolumn{1}{l|}{0.60}     &        \textbf{0.72}                  & \multicolumn{1}{l|}{\textbf{1.00}}   & \multicolumn{1}{l|}{0.52}    & \multicolumn{1}{l|}{\textbf{0.63}}     &       \textbf{0.72}                  & \multicolumn{1}{l|}{\textbf{0.98}}   & \multicolumn{1}{l|}{0.56}   & \multicolumn{1}{l|}{\textbf{0.61}}     &            \textbf{0.72}             &              \textbf{0.79}                              \\ \Xhline{1.5pt}   
\end{tabular}%
}
\end{table}
% \vspace{-1em}
\subsubsection{Results. }
We compared our approach performance with the AI4PAIN baseline methods with video modality. A Gaussian SVM (with an RBF kernel) was trained for the video-only. Features were extracted using the Py-Feat facial expression analysis toolbox \cite{pyfeat}. We further compare with other works \cite{ai4pain, painvit} which conducted their experiment on the AI4Pain dataset. Simple ANN with Majority Voting \cite{ai4pain} uses a straightforward neural network architecture to individually predict each frame in a video and then determine the final label through a majority vote. LSTM-based approach by \cite{ai4pain}, contained two LSTM layers with 32 and 16 units and trained with extracted VGG19 features from cropped-face videos. Twins-PainViT \cite{painvit} leverage ViT backbone to classify pain level for video modality. Results are shown in Table \ref{tab:1} and Table \ref{tab:3} for the Validation set and Test set, respectively. Our approach distinctly excels other baselines and other methods for precision, recall, and accuracy evaluation. The significant accuracy gap between the validation and test sets can be explained by the imbalance in the number of 'No Pain' samples, which our model evaluated most successfully on. We observed that No Pain videos account for 50\% in the validation set (including No Pain, Rest, and baseline stages) and around 6\% in unseen labeled test data (human evaluation). Additionally, we present our experiment results with different classifiers in Table \ref{tab:2}, showing that our designed skip MLP with ConvTrans achieved the highest performance.

% \usepackage{graphicx}
% \usepackage{graphicx}
% \usepackage{graphicx}

% \usepackage{multirow}
% \usepackage{graphicx}
% {\renewcommand{\arraystretch}{1.5}%

% {\renewcommand{\arraystretch}{1.5}%

% {\renewcommand{\arraystretch}{1.5}%

\section{Conclusion}
In this study, we presented a Transformer-based framework designed for automatic video pain assessment. Our model integrates a powerful Masked Auto Encoder to extract spatio-temporal features and utilizes a Residual Convolutional Transformer for predicting pain levels, enabling the comprehensive capture of facial attributes. We highlight the benefits of employing advanced transformer position encoding techniques, which enhance the positioning and embedding of time series data. 
% By effectively addressing the limitations of traditional self-reporting and observational methods, our approach significantly enhances pain recognition accuracy. 
Quantitative results show that our model surpasses other approaches to validation and test sets. 

\subsubsection{Ethical Impact Statement. }
This research employed the AI4PAIN dataset \cite{fernandez2023multimodal}, provided by the challenge organizers, to assess the proposed methods. Participants confirmed they had no history of neurological or psychiatric disorders, unstable medical conditions, chronic pain, or regular medication use at the time of the experiment. Before beginning the study, all participants received a detailed explanation of the experimental procedures, and written informed consent was obtained. The protocols involving human participants were reviewed and approved by the University of Canberra’s Human Ethics Committee (approval number: 11837).

The framework developed in this study aims to provide a reliable system for continuous pain monitoring while minimizing subjective human bias. Nevertheless, integrating this framework into real-world clinical environments poses challenges requiring additional testing and validation through comprehensive clinical trials. Furthermore, the facial image in this research is an artistic representation and does not correspond to any actual person.

\subsubsection{Acknowledgements.} This work was supported by the National Research Foundation of Korea (NRF) grant funded by the Korea government (MSIT) (RS-2023-00219107) and the Institute of Information \& communications Technology Planning \& Evaluation (IITP) under the Artificial Intelligence Convergence Innovation Human Resources Development (IITP- 2023-RS-2023-00256629) grant funded by the Korean government (MSIT).

\nocite{*}
\bibliographystyle{abbrv}
\bibliography{annot}

\begin{thebibliography}{10}

\bibitem{ganloss}
M.~Arjovsky, S.~Chintala, and L.~Bottou.
\newblock Wasserstein generative adversarial networks.
\newblock In {\em International conference on machine learning}, pages 214--223. PMLR, 2017.

\bibitem{ref6}
G.~Bargshady, X.~Zhou, R.~C. Deo, J.~Soar, F.~Whittaker, and H.~Wang.
\newblock Enhanced deep learning algorithm development to detect pain intensity from facial expression images.
\newblock {\em Expert systems with applications}, 149:113305, 2020.

\bibitem{Marlin}
Z.~Cai, S.~Ghosh, K.~Stefanov, A.~Dhall, J.~Cai, H.~Rezatofighi, R.~Haffari, and M.~Hayat.
\newblock Marlin: Masked autoencoder for facial video representation learning.
\newblock In {\em Proceedings of the IEEE/CVF conference on computer vision and pattern recognition}, pages 1493--1504, 2023.

\bibitem{vggface}
Q.~Cao, L.~Shen, W.~Xie, O.~M. Parkhi, and A.~Zisserman.
\newblock Vggface2: A dataset for recognising faces across pose and age.
\newblock In {\em 2018 13th IEEE international conference on automatic face \& gesture recognition (FG 2018)}, pages 67--74. IEEE, 2018.

\bibitem{pyfeat}
J.~H. Cheong, E.~Jolly, T.~Xie, S.~Byrne, M.~Kenney, and L.~J. Chang.
\newblock Py-feat: Python facial expression analysis toolbox.
\newblock {\em Affective Science}, 4(4):781--796, 2023.

\bibitem{vit}
A.~Dosovitskiy, L.~Beyer, A.~Kolesnikov, D.~Weissenborn, X.~Zhai, T.~Unterthiner, M.~Dehghani, M.~Minderer, G.~Heigold, S.~Gelly, et~al.
\newblock An image is worth 16x16 words: Transformers for image recognition at scale.
\newblock {\em arXiv preprint arXiv:2010.11929}, 2020.

\bibitem{fernandez2023multimodal}
R.~Fernandez~Rojas, N.~Hirachan, N.~Brown, G.~Waddington, L.~Murtagh, B.~Seymour, and R.~Goecke.
\newblock Multimodal physiological sensing for the assessment of acute pain.
\newblock {\em Frontiers in Pain Research}, 4:1150264, 2023.

\bibitem{convTrans}
N.~M. Foumani, C.~W. Tan, G.~I. Webb, and M.~Salehi.
\newblock Improving position encoding of transformers for multivariate time series classification.
\newblock {\em Data Mining and Knowledge Discovery}, 38(1):22--48, 2024.

\bibitem{ref7}
S.~Gkikas, N.~S. Tachos, S.~Andreadis, V.~C. Pezoulas, D.~Zaridis, G.~Gkois, A.~Matonaki, T.~G. Stavropoulos, and D.~I. Fotiadis.
\newblock Multimodal automatic assessment of acute pain through facial videos and heart rate signals utilizing transformer-based architectures.
\newblock {\em Frontiers in Pain Research}, 5:1372814, 2024.

\bibitem{painvit}
S.~Gkikas and M.~Tsiknakis.
\newblock Twins-painvit: Towards a modality-agnostic vision transformer framework for multimodal automatic pain assessment using facial videos and fnirs.
\newblock {\em arXiv preprint arXiv:2407.19809}, 2024.

\bibitem{lstm}
S.~Hochreiter and J.~Schmidhuber.
\newblock Long short-term memory.
\newblock {\em Neural computation}, 9(8):1735--1780, 1997.

\bibitem{ref4}
N.~Kalischek, P.~Thiam, P.~Bellmann, and F.~Schwenker.
\newblock Deep domain adaptation for facial expression analysis.
\newblock In {\em 2019 8th International Conference on Affective Computing and Intelligent Interaction Workshops and Demos (ACIIW)}, pages 317--323. IEEE, 2019.

\bibitem{ref1}
G.~Lu, H.~Chen, J.~Wei, X.~Li, X.~Zheng, H.~Leng, Y.~Lou, and J.~Yan.
\newblock Video-based neonatal pain expression recognition with cross-stream attention.
\newblock {\em Multimedia Tools and Applications}, 83(2):4667--4690, 2024.

\bibitem{ai4pain}
P.~Prajod, D.~Schiller, D.~W. Don, and E.~Andr{\'e}.
\newblock Faces of experimental pain: Transferability of deep learned heat pain features to electrical pain.
\newblock {\em arXiv preprint arXiv:2406.11808}, 2024.

\bibitem{ref3}
P.~Rodriguez, G.~Cucurull, J.~Gonz{\`a}lez, J.~M. Gonfaus, K.~Nasrollahi, T.~B. Moeslund, and F.~X. Roca.
\newblock Deep pain: Exploiting long short-term memory networks for facial expression classification.
\newblock {\em IEEE transactions on cybernetics}, 52(5):3314--3324, 2017.

\bibitem{p2}
J.~Strong, A.~Unruh, A.~Wright, and G.~Baxter.
\newblock Pain: a textbook for therapists.
\newblock 2002.

\bibitem{ref5}
M.~Tavakolian and A.~Hadid.
\newblock A spatiotemporal convolutional neural network for automatic pain intensity estimation from facial dynamics.
\newblock {\em International Journal of Computer Vision}, 127:1413--1425, 2019.

\bibitem{trans}
A.~Vaswani, N.~Shazeer, N.~Parmar, J.~Uszkoreit, L.~Jones, A.~N. Gomez, {\L}.~Kaiser, and I.~Polosukhin.
\newblock Attention is all you need.
\newblock {\em Advances in neural information processing systems}, 30, 2017.

\bibitem{face}
J.~Wang, Y.~Liu, Y.~Hu, H.~Shi, and T.~Mei.
\newblock Facex-zoo: A pytorch toolbox for face recognition.
\newblock In {\em Proceedings of the 29th ACM international conference on Multimedia}, pages 3779--3782, 2021.

\bibitem{p1}
S.~M. Zwakhalen, J.~P. Hamers, H.~H. Abu-Saad, and M.~P. Berger.
\newblock Pain in elderly people with severe dementia: a systematic review of behavioural pain assessment tools.
\newblock {\em BMC geriatrics}, 6:1--15, 2006.

\end{thebibliography}

\end{document}